# Increasing Melanoma Diagnostic Confidence: Forcing the Convolutional Network to Learn from the Lesion

Norsang Lama, R. Joe Stanley, Anand Nambisan, Akanksha Maurya, Jason Hagerty, and William V. Stoecker

*Abstract* — **Deep learning implemented with convolutional network architectures can exceed specialists' diagnostic accuracy. However, whole-image deep learning trained on a given dataset may not generalize to other datasets. The problem arises because extra-lesional features—ruler marks, ink marks, and other melanoma correlates—may serve as information leaks. These extra-lesional features, discoverable by heat maps, degrade melanoma diagnostic performance and cause techniques learned on one data set to fail to generalize. We propose a novel technique to improve melanoma recognition by an EfficientNet model. The model trains the network to detect the lesion and learn features from the detected lesion. A generalizable elliptical segmentation model for lesions was developed, with an ellipse enclosing a lesion and the ellipse enclosed by an extended rectangle (bounding box). The minimal bounding box was extended by 20% to allow some background around the lesion. The publicly available International Skin Imaging Collaboration (ISIC) 2020 skin lesion image dataset was used to evaluate the effectiveness of the proposed method. Our test results show that the proposed method improved diagnostic accuracy by increasing the mean area under receiver operating characteristic curve (mean AUC) score from 0.9 to 0.922. Additionally, correctly diagnosed scores are also improved, providing better separation of scores, thereby increasing melanoma diagnostic confidence. The proposed lesion-focused convolutional technique warrants further study.**

*Index Terms* — *melanoma, dermoscopy, deep learning, image classification, skin lesion*

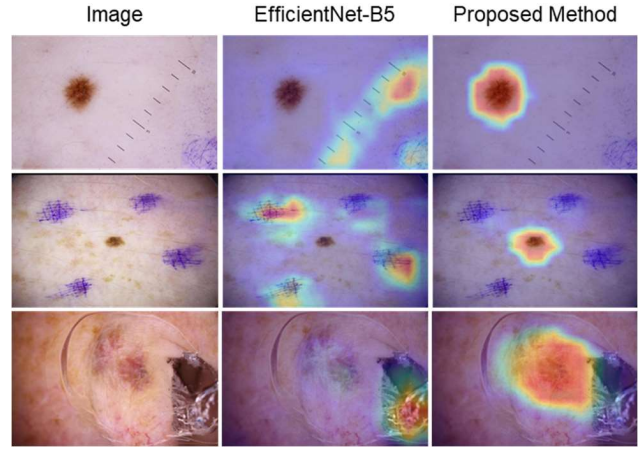

Fig. 1. CAM heatmap visualizations of an EfficientNet-B5 model and the proposed method for melanoma classification.

## I. INTRODUCTION

AN estimated 186,680 new cases of invasive melanoma (including 89,070 in-situ melanomas) are expected to be diagnosed in 2023 in the United States [1]. Dermoscopy is an imaging adjunct technique for early skin cancer detection, improving diagnostic accuracy compared to visual inspection by a domain expert [2]–[4].

Computer vision techniques have improved appreciably in recent years [5]–[12] and have been successfully applied to many medical imaging problems [13]–[20]. In the skin cancer domain, deep learning techniques combined with dermoscopy have higher diagnostic accuracy than experienced dermatologists [13], [21]–[24]. Pathan *et al*. published a recent review detailing both handcrafted and deep learning (DL) techniques for computer-aided diagnosis of skin lesions [25]. Recent studies show that the fusion of deep learning and handcrafted features can improve accuracy in skin cancer diagnosis [24], [26]–[30].

Although convolution neural network (CNN) methods have achieved higher diagnostic accuracy in skin lesion classification, the heatmap visualizations of CNNs have shown that they do not always learn the features from a lesion region in the image; rather from the artifacts present in the image, such as ruler marks, ink marks, stickers, and skin backgrounds. These non-lesional features may serve as information leaks and might potentially cause poor generalization when applied to new test data that are different than training data. Thus, in this study, we propose a novel deep learning method that forces CNN, in particular an EfficientNet-B5 model, to learn the features from the important lesion region in the image during the training. The class activation map (CAM) visualizations in Figure 1 shows that the proposed method prevents CNN model focusing on the artifacts. Furthermore, the test results show the proposed method improves the melanoma classification performance and predicts the classification score with a higher diagnostic confidence.

Date submitted 15 May 2023. This work was supported in part by the National Institutes of Health (NIH) under Grant SBIR R43 CA153927- 01 and CA101639-02A2 of the National Institutes of Health (NIH) and Grant OAC-1919789 of the National Science Foundation (NSF). Its contents are solely the responsibility of the authors and do not necessarily represent the official views of the NIH or NSF. (Corresponding author: R. Joe Stanley)

N. Lama, A. Maurya, A. Nambisan, and R. Joe Stanley are with Missouri University of Science & Technology, Rolla, MO 65209 USA (email: nlbft@mst.edu; amvq5@umsystem.edu; akn36d@mst.edu; stanleyj@mst.edu).

J. Hagerty, and W. V. Stoecker are with S&A Technologies, Rolla, MO, 65401 USA (email: hagerty.jason@gmail.com; wvs@mst.edu).

## II. MATERIALS AND METHODS

### A. Image Datasets

In this study, we used a publicly available ISIC2020 [31] melanoma classification dataset. The dataset has 33,126 skin lesion dermoscopic images of two categories – benign and melanoma. Some of the images have duplicates; we created a curated set of 32701 images after removing the duplicates. The dataset is highly imbalanced with only 581 (1.78%) of total images belongs melanoma category The images have varying resolutions from 480×640 to 4000×6000. Some of the examples are shown in Figure 2. The non-square images were zero padded and resized to 512x512 using a bilinear interpolation.

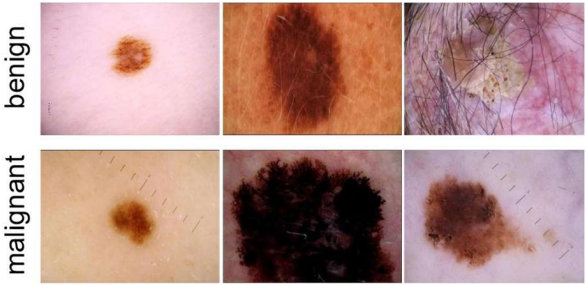

Fig. 2. Skin lesion dermoscopy images with ground truth classification labels in the ISIC 2020 skin lesion dataset. The top row shows benign lesions, and the bottom row shows malignant ones.

### B. Data Augmentation

In this study, we applied the data augmentation during the training of convolutional neural network. It increases the variation in training images by randomly applying various image transformations, which eventually helps the model to generalize better. The image transformations used in this study are as follows:

- Transpose
- Horizontal or Vertical Flip
- Height or width shift with a range of (-0.15, +0.15)
- Rotation with range between +90° to -90°
- Zoom with a range of (0.85, 1.15)
- Brightness with a range of (0.85, 1.15)
- Contrast with a range of (0.85, 1.15)
- Hue with a range of (0.85, 1.15)
- Saturation with a range of (0.85, 1.15)
- CLAHE histogram equalization
- Gaussian Noise
- Motion Blur
- Median Blur
- Gaussian Blur

Furthermore, the image pixel values were rescaled between 0 and 1 and normalized using the ImageNet [32] parameters.

### C. Proposed Method

The overall flow diagram of the proposed method is shown in Figure 3. It uses a pretrained EfficientNet [12] model as a convolutional neural network (CNN) architecture to classify the skin lesions. It incorporates a novel attention mechanism to force the model to focus more on the lesion region of an image. The proposed attention mechanism, first, computes the class activation map (CAM) [33] to identify the image regions most relevant to the specific class (melanoma in our case) and then uses it with an elliptical lesion mask to compute the attention loss, $L_A$. The attention loss, $L_A$, is combined with the classification loss, $L_C$, to create the composite loss $L_T$. Finally, the convolutional neural network is trained using this composite loss so that the network emphasizes more on the lesion region in the image rather than the background. For a given image, let $f_k(x, y)$ represent an activation of a unit $k$ in the last convolution layer at a spatial location $(x, y)$. The CAM for class $c$ is given in Equation 1.

$$M_c(x, y) = \sum_k w_k^c f_k(x, y) \tag{1}$$

Where $w_k^c$ is the weight corresponding to class $c$ for the unit $k$.

To generate an elliptical lesion mask, $M_E$, we use an extended bounding box that encloses the skin lesion. The bounding box of the lesion images are auto generated using a separate lesion detection model, which is a ResNet-50 [9] model trained on the ISIC 2018 [34] lesion segmentation dataset and predicts the bounding box coordinates $(x_{min}, y_{min}, x_{max}, y_{max})$ of a lesion. The bounding box is extended by 20% area to allow some background around the lesion. The elliptical mask $M_E$ is resized to same size as that of $M_C$ to compute the attention loss. Here, $M_E$ is a binary mask with a pixel value of 0 or 1. Thus, the data range of $M_C$ is also rescaled between 0 and 1 using a division by maximum value. For $N$ training images, the attention loss using the Jaccard method is given in Equation 2.

$$L_A = 1 - \frac{\sum_N M_C M_E + 1}{\sum_N M_C + M_E + 1} \tag{2}$$

Equation 3 shows the classification loss $L_C$ computed using a cross-entropy method between a sigmoid output of a fully connected (FC) layer, $S_{FC}$, and a given ground truth label $Y$.

$$L_C = BCE(S_{FC}, Y) \tag{3}$$

Equation 4 shows the total composite loss that was used to train the network.

$$L_T = (1 - \lambda)L_C + \lambda L_A \tag{4}$$

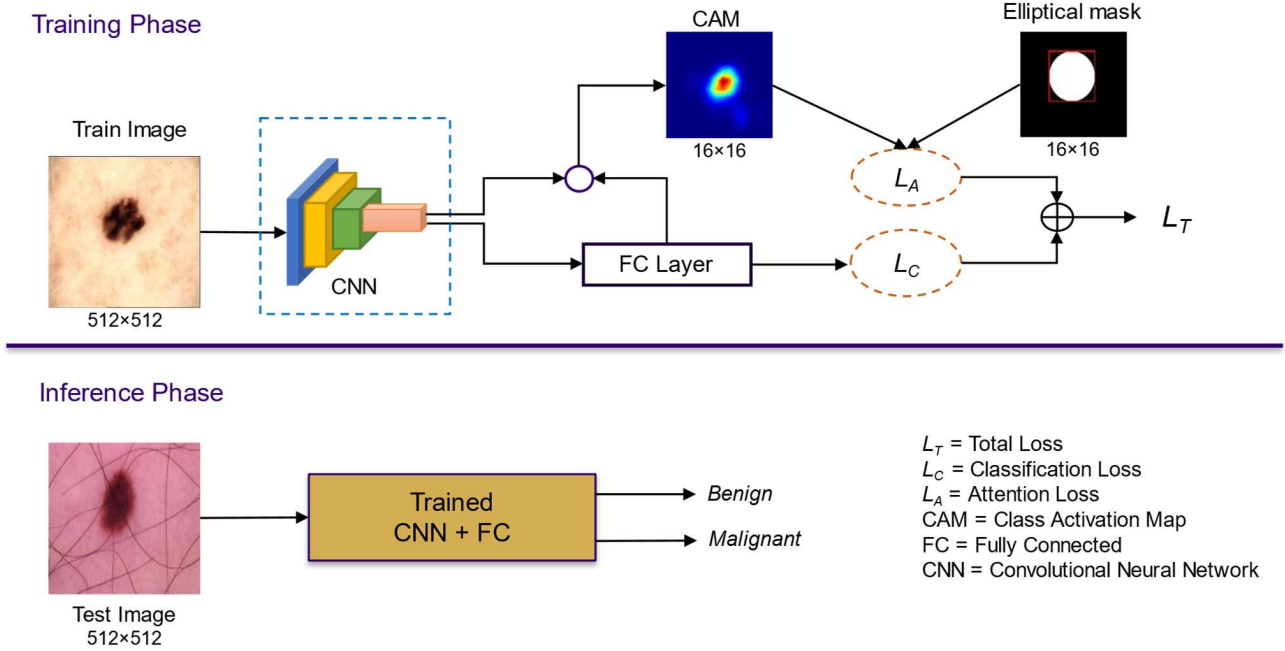

Fig. 3. The overall flow diagram of a proposed melanoma classification method. During training, the attention mechanism computes the class activation map (CAM) using the feature map after the last convolutional layer, which is further used to compute the attention loss $L_A$. The classification loss $L_C$ is computed using an output from FC layer and combined to create a composite (total) loss $L_T$.

Where $\lambda$ is the loss weight factor such that $0 < \lambda < 1$, and was optimized empirically to 0.66.

During the inference, the trained CNN model with the fully connected (FC) layer outputs the sigmoid score with a value between 0 and 1, where the score closer to 0 indicates the lesion being benign and the score closer to 1 indicates lesion being melanoma.

### D. Training Details

All models were built using a PyTorch framework in Python 3 and trained using a single 32GB Nvidia V100 graphics card. The network was trained for 30 epochs using a batch size of 6, a constant learning rate of 0.0001, and the stochastic gradient descent (SGD) optimization algorithm. The loss functions were weighted binary cross entropy for a classification loss and Jaccard loss for an attention loss. To reduce overfitting of a deep neural network model, we used data augmentation (see details in section II.B), a dropout layer, and an early stopping technique. The dropout probability of 0.5 was selected for the dropout layer, which was placed before a FC layer. For the early stopping criterion, we used a patience of 5 epochs to stop the model from overtraining.

## III.   EXPERIMENTAL RESULTS

To evaluate the performance of the proposed method, we trained an Efficient-B5 model with our proposed attention mechanism (AM) using 5-fold cross validation. The 32,701 images from the curated ISIC dataset were randomly split into 5 folds with a class label-based stratification. We used the area under the receiver operating characteristic curve (AUC) to measure the classification performance of the proposed model.

TABLE I PEFORMANCE COMPARISON OF THE PROPOSED METHOD AGAINST THE BASELINE MODEL ON ISIC2020 DATASET

|  | AUC | | |
|---|---|---|---|
|  | Median | Mean | Standard Deviation |
| EfficientNet (baseline) | 0.897 | 0.9 | 0.0106 |
| EfficientNet + AM | 0.931 | 0.922 | 0.0167 |

Table I shows the performance comparison of the proposed method against the baseline model. The baseline model is the Efficient-B5 model without the attention mechanism. The proposed method improved the mean cross-validated AUC of 0.9 to 0.922. In Figures 4, we show the class activation map (CAM) of the proposed method on the test melanoma images. Although the baseline model has a prediction score greater than 0.5 in all three cases, CAM shows the model focuses on the outer regions (example, ruler marks) rather than the lesion region. In contrast, the proposed method focuses mostly inside the lesion bounding box. Also, the prediction scores of 0.818 vs. 0.745, 0.739 vs. 0.703 and 0.90 vs. 0.701 from the proposed model against the baseline model shows that the proposed model is more confident of classifying the melanoma lesions as a melanoma.

Similarly, Figure 5 shows the overlays of CAM on the benign images from both proposed and baseline models. The proposed model focuses within the lesion region to extract the important information to classify the sample correctly. Conversely, the baseline model focuses on the image corner or the background regions even though it is correctly predicting the lesions as a benign lesion. Also, the prediction scores are reduced from 0.448 to 0.099, 0.279 to 0.092 and 0.037 to 0.007, showing the improved confidence on its classification score.

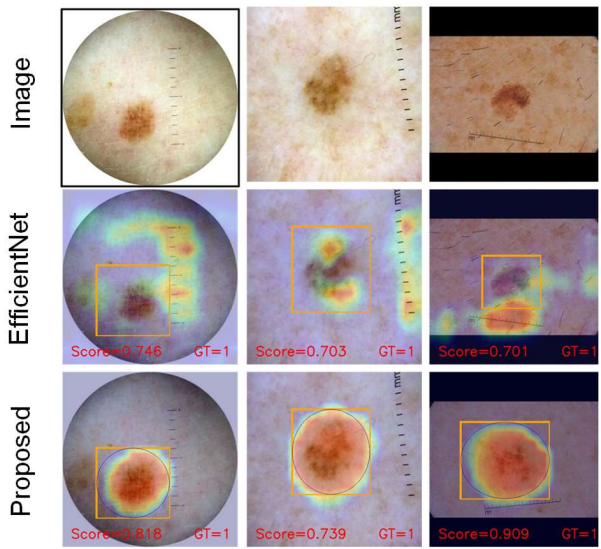

Fig. 4. Overlays of class activation map (CAM) on the test melanoma lesion images. The bounding box shows lesion location. The CAM shows the proposed method focuses within the lesion region. The scores and GT in RED show the proposed method is more confident of classifying the melanoma lesions as a melanoma.

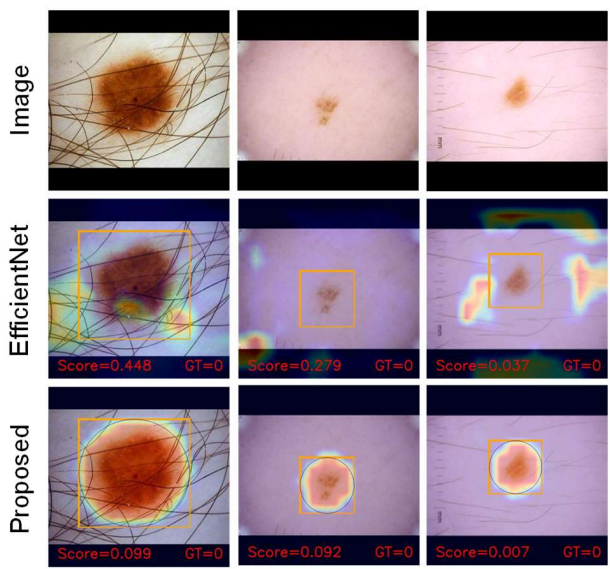

Fig. 5. Overlays of the class activation map (CAM) on the test benign lesion images. The bounding box shows lesion location. The CAM shows the proposed method focuses within the lesion region. The scores and GT in RED show the proposed method is more confident of classifying the benign lesions as benign.

## IV. DISCUSSION

In this study, we demonstrated that our proposed lesion-focused deep learning method not only improves the melanoma classification performance, but also increases melanoma diagnostic confidence in dermoscopic skin lesion images. As accuracy is not a very useful metric for a binary classification problem in a highly imbalanced dataset scenario, we used an area under ROC curve (AUC) to evaluate the classification performance of the proposed method.

In recent ISIC skin lesion classification challenges, DL methods using CNN architectures such as ResNet [9], ResNext[35], SEResNext [36], DenseNet [37] and EfficientNet [12] have dominated the submission leaderboards [38], [39]. Although CNN methods have outperformed the traditional handcrafted features methods in visual recognition tasks, it is little known how they perform very well. Various visualization techniques, including saliency maps [40], CAM [33], Grad-CAM [41], Grad-CAM++ [42], and Score-CAM [43], have been devised to observe the specific regions in an image that played a significant role in the classification of a particular class by convolutional neural networks.

Cassidy et al. [44] recently analyzed the ISIC image datasets using various convolution neural networks. In their study, a Grad-CAM visualization showed the CNN models primarily focus on non-lesion regions in an image such as ruler marks, ink marks, stickers, and skin background. Furthermore, we noticed the similar behavior of the CNN model in this investigation. Despite CNNs focusing on non-lesion regions, they still manage to make the correct predictions as shown in Figures 4 and 5. This situation is undesirable since the presence of such artifacts in the training data can lead to information leakage, potentially causing the trained model to perform poorly when applied to new test images from different distributions than the training data. Thus, the CNN methods that focuses on the lesion regions are warranted to develop a better generalized model.

Our experimental results showed that the proposed attention mechanism forces the CNN model to learn from the important lesion regions in an image. The CNN model trained with the proposed attention mechanism achieves an improved classification performance compared to the plain (no attention component) CNN model. Also, the model with attention predicts the classifications scores with higher confidence than the baseline model. As the proposed CNN model mainly relies on the lesion region in the image to make a final classification prediction, such models can perform well even when clinical artifacts are not present in the test images. However, the generalization capability of the proposed method is not investigated in the current study, as such an investigation can be performed only when new test data with a different distribution than the training data are available in the future.

## V. CONCLUSION

In this study, we propose a novel deep learning technique to force a convolutional neural network (CNN) to learn from an important lesion region in dermoscopic skin lesion images. The proposed method employs a new attention mechanism that uses a class activation map and an elliptical lesion mask to compute an attention loss. The attention loss is combined with the classification loss to train the convolutional neural network. The CNN model trained with the combined loss improved the melanoma classification performance. The class activation map showed that the CNN model with the proposed method makes an accurate melanoma prediction by learning features within a lesion rather than non-lesion regions in the skin background.

## REFERNCES

[1] R. L. Siegel, K. D. Miller, N. S. Wagle, and A. Jemal, "Cancer statistics, 2023," *CA Cancer J Clin*, vol. 73, no. 1, pp. 17–48, 2023, doi: https://doi.org/10.3322/caac.21763.

[2] H. Pehamberger, M. Binder, A. Steiner, and K. Wolff, "In vivo epiluminescence microscopy: Improvement of early diagnosis of melanoma," *Journal of Investigative Dermatology*, vol. 100, no. 3 SUPPL., pp. S356–S362, 1993, doi: 10.1038/jid.1993.63.

[3] H. P. Soyer, G. Argenziano, R. Talamini, and S. Chimenti, "Is Dermoscopy Useful for the Diagnosis of Melanoma?," *Arch Dermatol*, vol. 137, no. 10, pp. 1361–1363, Oct. 2001, doi: 10.1001/archderm.137.10.1361.

[4] R. P. Braun, H. S. Rabinovitz, M. Oliviero, A. W. Kopf, and J. H. Saurat, "Pattern analysis: a two-step procedure for the dermoscopic diagnosis of melanoma," *Clin Dermatol*, vol. 20, no. 3, pp. 236–239, May 2002, doi: 10.1016/S0738-081X(02)00216-X.

[5] A. Krizhevsky, I. Sutskever, and G. Hinton, "ImageNet Classification with Deep Convolutional Neural Networks," in *Advances in Neural Information and Processing Systems (NIPS)*, 2012, pp. 1097–1105.

[6] C. Szegedy, V. Vanhoucke, S. Ioffe, J. Shlens, and Z. Wojna, "Rethinking the inception architecture for computer vision," in *Proceedings of the IEEE conference on computer vision and pattern recognition*, 2016, pp. 2818–2826.

[7] K. Simonyan and A. Zisserman, "Very deep convolutional networks for large-scale image recognition," *arXiv preprint arXiv:1409.1556*, 2014.

[8] I. Goodfellow *et al.*, "Generative adversarial networks," *Commun ACM*, vol. 63, no. 11, pp. 139–144, 2020.

[9] K. He, X. Zhang, S. Ren, and J. Sun, "Deep residual learning for image recognition," in *Proceedings of the IEEE conference on computer vision and pattern recognition*, 2016, pp. 770–778.

[10] A. Dosovitskiy *et al.*, "An image is worth 16x16 words: Transformers for image recognition at scale," *arXiv preprint arXiv:2010.11929*, 2020.

[11] O. Ronneberger, P. Fischer, and T. Brox, "U-Net: Convolutional Networks for Biomedical Image Segmentation." [Online]. Available: http://lmb.informatik.uni-freiburg.de/

[12] M. Tan and Q. Le, "Efficientnet: Rethinking model scaling for convolutional neural networks," in *International conference on machine learning*, 2019, pp. 6105–6114.

[13] A. Esteva *et al.*, "Dermatologist-level classification of skin cancer with deep neural networks," *Nature*, vol. 542, no. 7639, pp. 115–118, 2017, doi: 10.1038/nature21056.

[14] V. Gulshan *et al.*, "Development and validation of a deep learning algorithm for detection of diabetic retinopathy in retinal fundus photographs," *JAMA*, vol. 316, no. 22, pp. 2402–2410, 2016.

[15] S. Sornapudi *et al.*, "Deep learning nuclei detection in digitized histology images by superpixels," *J Pathol Inform*, vol. 9, no. 1, p. 5, 2018.

[16] G. Litjens *et al.*, "A survey on deep learning in medical image analysis," *Med Image Anal*, vol. 42, pp. 60–88, 2017, doi: https://doi.org/10.1016/j.media.2017.07.005.

[17] A. K. Nambisan *et al.*, "Deep learning-based dot and globule segmentation with pixel and blob-based metrics for evaluation," *Intelligent Systems with Applications*, vol. 16, p. 200126, 2022, doi: https://doi.org/10.1016/j.iswa.2022.200126.

[18] N. Lama, J. Hagerty, A. Nambisan, R. J. Stanley, and W. Van Stoecker, "Skin Lesion Segmentation in Dermoscopic Images with Noisy Data," *J Digit Imaging*, 2023, doi: 10.1007/s10278-023-00819-8.

[19] A. Maurya *et al.*, "A deep learning approach to detect blood vessels in basal cell carcinoma," *Skin Research and Technology*, vol. 28, no. 4, pp. 571–576, 2022.

[20] N. Lama *et al.*, "ChimeraNet: U-Net for Hair Detection in Dermoscopic Skin Lesion Images," *J Digit Imaging*, no. 0123456789, 2022, doi: 10.1007/s10278-022-00740-6.

[21] L. K. Ferris *et al.*, "Computer-aided classification of melanocytic lesions using dermoscopic images," *J Am Acad Dermatol*, vol. 73, no. 5, pp. 769–776, Nov. 2015, doi: 10.1016/J.JAAD.2015.07.028.

[22] M. A. Marchetti *et al.*, "Results of the 2016 International Skin Imaging Collaboration International Symposium on Biomedical Imaging challenge: Comparison of the accuracy of computer algorithms to dermatologists for the diagnosis of melanoma from dermoscopic images," *J Am Acad Dermatol*, vol. 78, no. 2, pp. 270-277.e1, Feb. 2018, doi: 10.1016/j.jaad.2017.08.016.

[23] H. A. Haenssle *et al.*, "Man against machine: diagnostic performance of a deep learning convolutional neural network for dermoscopic melanoma recognition in comparison to 58 dermatologists," *Annals of Oncology*, vol. 29, no. 8, pp. 1836–1842, 2018, doi: https://doi.org/10.1093/annonc/mdy166.

[24] N. C. F. Codella *et al.*, "Deep Learning Ensembles for Melanoma Recognition in Dermoscopy Images," *IBM J. Res. Dev.*, vol. 61, no. 4–5, pp. 5:1–5:15, Jul. 2017, doi: 10.1147/JRD.2017.2708299.

[25] S. Pathan, K. G. Prabhu, and P. C. Siddalingaswamy, "Techniques and algorithms for computer aided diagnosis of pigmented skin lesions—A review," *Biomed Signal Process Control*, vol. 39, pp. 237–262, Jan. 2018, doi: 10.1016/J.BSPC.2017.07.010.

[26] T. Majtner, S. Yildirim-Yayilgan, and J. Y. Hardeberg, "Combining deep learning and hand-crafted features for skin lesion classification," *2016 6th International Conference on Image Processing*

*Theory, Tools and Applications, IPTA 2016*, 2017, doi: 10.1109/IPTA.2016.7821017.

[27] N. Codella, J. Cai, M. Abedini, R. Garnavi, A. Halpern, and J. R. Smith, "Deep Learning, Sparse Coding, and SVM for Melanoma Recognition in Dermoscopy Images BT - Machine Learning in Medical Imaging," L. Zhou, L. Wang, Q. Wang, and Y. Shi, Eds., Cham: Springer International Publishing, 2015, pp. 118–126.

[28] I. González-Díaz, "DermaKNet: Incorporating the Knowledge of Dermatologists to Convolutional Neural Networks for Skin Lesion Diagnosis," *IEEE J Biomed Health Inform*, vol. 23, no. 2, pp. 547–559, 2019, doi: 10.1109/JBHI.2018.2806962.

[29] J. R. Hagerty *et al.*, "Deep Learning and Handcrafted Method Fusion: Higher Diagnostic Accuracy for Melanoma Dermoscopy Images," *IEEE J Biomed Health Inform*, vol. 23, no. 4, pp. 1385–1391, 2019, doi: 10.1109/JBHI.2019.2891049.

[30] A. K. Nambisan *et al.*, "Improving Automatic Melanoma Diagnosis Using Deep Learning-Based Segmentation of Irregular Networks," *Cancers (Basel)*, vol. 15, no. 4, 2023, doi: 10.3390/cancers15041259.

[31] V. Rotemberg *et al.*, "A patient-centric dataset of images and metadata for identifying melanomas using clinical context," *Sci Data*, vol. 8, no. 1, p. 34, 2021, doi: 10.1038/s41597-021-00815-z.

[32] J. Deng, W. Dong, R. Socher, L.-J. Li, K. Li, and L. Fei-Fei, "Imagenet: A large-scale hierarchical image database," in *2009 IEEE conference on computer vision and pattern recognition*, 2009, pp. 248–255.

[33] B. Zhou, A. Khosla, A. Lapedriza, A. Oliva, and A. Torralba, "Learning deep features for discriminative localization," in *Proceedings of the IEEE conference on computer vision and pattern recognition*, 2016, pp. 2921–2929.

[34] N. Codella *et al.*, "Skin Lesion Analysis Toward Melanoma Detection 2018: A Challenge Hosted by the International Skin Imaging Collaboration (ISIC)," pp. 1–12, 2019, [Online]. Available: http://arxiv.org/abs/1902.03368

[35] S. Xie, R. Girshick, P. Dollár, Z. Tu, and K. He, "Aggregated residual transformations for deep neural networks," in *Proceedings of the IEEE conference on computer vision and pattern recognition*, 2017, pp. 1492–1500.

[36] J. Hu, L. Shen, and G. Sun, "Squeeze-and-excitation networks," in *Proceedings of the IEEE conference on computer vision and pattern recognition*, 2018, pp. 7132–7141.

[37] G. Huang, Z. Liu, L. Van Der Maaten, and K. Q. Weinberger, "Densely connected convolutional networks," in *Proceedings of the IEEE conference on computer vision and pattern recognition*, 2017, pp. 4700–4708.

[38] International Skin Imaging Collaboration, "ISIC 2019 Leaderboards," May 14, 2019. https://challenge.isic-archive.com/leaderboards/2019/ (accessed May 13, 2023).

[39] Society for Imaging Informatics in Medicine (SIIM) and International Skin Imaging Collaboration (ISIC), "SIIM-ISIC Melanoma Classification Leaderboard," May 14, 2020. https://www.kaggle.com/competitions/siim-isic-melanoma-classification/leaderboard (accessed May 13, 2023).

[40] K. Simonyan, A. Vedaldi, and A. Zisserman, "Deep inside convolutional networks: Visualising image classification models and saliency maps," *arXiv preprint arXiv:1312.6034*, 2013.

[41] R. R. Selvaraju, M. Cogswell, A. Das, R. Vedantam, D. Parikh, and D. Batra, "Grad-cam: Visual explanations from deep networks via gradient-based localization," in *Proceedings of the IEEE international conference on computer vision*, 2017, pp. 618–626.

[42] A. Chattopadhay, A. Sarkar, P. Howlader, and V. N. Balasubramanian, "Grad-cam++: Generalized gradient-based visual explanations for deep convolutional networks," in *2018 IEEE winter conference on applications of computer vision (WACV)*, 2018, pp. 839–847.

[43] H. Wang *et al.*, "Score-CAM: Score-weighted visual explanations for convolutional neural networks," in *Proceedings of the IEEE/CVF conference on computer vision and pattern recognition workshops*, 2020, pp. 24–25.

[44] B. Cassidy, C. Kendrick, A. Brodzicki, J. Jaworek-Korjakowska, and M. H. Yap, "Analysis of the ISIC image datasets: Usage, benchmarks and recommendations," *Med Image Anal*, vol. 75, p. 102305, 2022, doi: https://doi.org/10.1016/j.media.2021.102305.